\documentclass[11pt,a4paper]{article}
\usepackage{acl2019}
\usepackage{times,latexsym}  
\usepackage{helvet}  
\usepackage{courier}  
\usepackage{url}  
\usepackage{graphicx}  
\usepackage{amsfonts}
\usepackage{amssymb}
\usepackage{amsthm}
\usepackage{amsmath}
\usepackage{algpseudocode}
\usepackage{algorithm}
\usepackage[T1]{fontenc}
\usepackage{xspace,mfirstuc,tabulary}

\newcommand{\C}{\ensuremath{\mathbf{C}}}

\newcommand{\HH}{\ensuremath{\mathbf{H}}}

\newcommand{\W}{\ensuremath{\mathbf{W}}}
\newcommand{\X}{\ensuremath{\mathbf{X}}}

\renewcommand{\b}{\ensuremath{\mathbf{b}}}
\renewcommand{\c}{\ensuremath{\mathbf{c}}}

\newcommand{\f}{\ensuremath{\mathbf{f}}}
\newcommand{\g}{\ensuremath{\mathbf{g}}}
\newcommand{\h}{\ensuremath{\mathbf{h}}}

\newcommand{\sss}{\ensuremath{\mathbf{s}}}  

\newcommand{\uu}{\ensuremath{\mathbf{u}}}
\newcommand{\vv}{\ensuremath{\mathbf{v}}}

\newcommand{\x}{\ensuremath{\mathbf{x}}}
\newcommand{\y}{\ensuremath{\mathbf{y}}}

\newcommand{\btheta}{\ensuremath{\boldsymbol{\theta}}}

\newcommand{\bbR}{\ensuremath{\mathbb{R}}}

\aclfinalcopy
\title{Towards Linear Time Neural Machine Translation with Capsule Networks}

\author{Mingxuan Wang$^1$ \ Jun xie$^2$ \
         Zhixing Tan$^2$ \  Jinsong Su$^3$ \ Deyi Xiong$^4$ \ Lei Li$^1$\\
        $^1$ByteDance AI Lab, Beijing, China\\
        {\tt \{wangmingxuan.89,lileilab\}@bytedance.com}\\
        $^2$Mobile Internet Group, Tencent Technology Co., Ltd\\
        $^3$Xiamen University, Xiamen, China\\
        $^4$Tianjin University, Tianjin, China\\
}

\begin{document}
%
\maketitle
\begin{abstract}
In this study, we  first investigate a novel capsule network
with dynamic routing for linear time Neural  Machine Translation (NMT), referred as
\textsc{CapsNMT}.
\textsc{CapsNMT}
uses an aggregation mechanism to map the source sentence into a matrix with pre-determined
size, and then applys a deep LSTM network to decode the
target sequence from the source representation.
Unlike the previous
work \cite{sutskever2014sequence} to store
the  source sentence with a passive and bottom-up way,
the dynamic routing policy encodes the source sentence with
an iterative process to
decide the credit attribution between nodes from lower and higher layers.
\textsc{CapsNMT} has two core properties:
it runs in time that is linear in the length of the sequences
and
provides a more flexible way to aggregate the part-whole information of the source
sentence.
On WMT14 English-German task and a larger WMT14 English-French task, \textsc{CapsNMT} achieves comparable results
with the Transformer system.
We also devise  new
hybrid architectures intended to combine the strength of \textsc{CapsNMT} and the RNMT model.
Our hybrid models obtain
state-of-the-arts results on both benchmark
datasets.
To the best of our knowledge, this is the first work that capsule networks have been empirically investigated
for  sequence to sequence problems.\footnote{The work is partially done when the first author worked at Tencent.}
\end{abstract}
\section{Introduction}
Neural Machine Translation (NMT)
is an end-to-end learning approach to machine  translation
which has recently shown
promising results on multiple language
pairs~\cite{luong2015effective,shen2015minimum,wu2016google,DBLP:journals/corr/GehringAGYD17,kalchbrenner2016neural,DBLP:journals/corr/SennrichHB15,vaswani2017attention}.
Unlike conventional Statistical Machine
Translation (SMT) systems~\cite{koehn2003statistical,chiang2005hierarchical} which consist of multiple
separately tuned components,
NMT aims at building upon a single and large neural
network to directly map input
text to associated output text ~\cite{sutskever2014sequence}.

In general, there are several research lines of NMT architectures, among which
the Enc-Dec NMT ~\cite{sutskever2014sequence}
and the Enc-Dec Att NMT  are of typical representation
~\cite{RNNsearch,wu2016google,vaswani2017attention}.
The Enc-Dec  represents the
source inputs with a fixed dimensional vector and
the target sequence is generated from this vector
word by word. The Enc-Dec, however, does not
preserve the source sequence resolution, a feature which aggravates learning for long sequences.
This results in
the computational complexity of decoding process being $\mathcal{O}(|S|+|T|)$, with $|S|$
denoting the source sentence length and  $|T|$ denoting the target sentence length.
The  Enc-Dec Att  preserves
the resolution of the source sentence which frees the neural model
from having to squash all the information into a fixed represention, but at a cost of a quadratic running time.
Due to the attention mechanism, the computational complexity of decoding process is $\mathcal{O}(|S|\times|T|)$.
 This drawbacks
grow more severe as the length of the sequences increases.

Currently, most work focused on the
Enc-Dec Att, while the Enc-Dec
paradigm is less emphasized on despite its advantage of linear-time decoding~\cite{kalchbrenner2016neural}.
The linear-time approach is appealing, however, the performance lags behind the Enc-Dec Att.
One potential issue is that the Enc-Dec needs to be able to
compress all the necessary information of the input source sentence into
context vectors which are fixed during decoding.
Therefore,
a natural question was raised, \emph{Will  carefully designed
aggregation operations help the Enc-Dec paradigm to achieve the best performance?}

In recent promising work of capsule network, a dynamic routing policy is proposed and proven to
effective~\cite{sabour2017dynamic,zhao2018investigating,gong2018information}.
Following a similar spirit to use this technique, we present \textsc{CapsNMT},
which is characterized by
capsule encoder to address the drawbacks of the conventional
linear-time approaches.
The capsule encoder processes the attractive
potential to address the aggregation issue, and
then introduces an iterative routing policy to decide the
credit attribution
between nodes from lower (child) and higher (parent) layers.
Three strategies are also proposed to stabilize the dynamic routing process.
We empirically verify  \textsc{CapsNMT}
on WMT14 English-German task and a larger WMT14 English-French task.
\textsc{CapsNMT} achieves comparable results
with the state-of-the-art Transformer systems.
Our contributions can be summrized as follows:
\begin{itemize}
  \item
  We propose a sophisticated designed linear-time
  \textsc{CapsNMT} which achieved
  comparable results with the Transformer framework.
  To the best of our knowledge, \textsc{CapsNMT} is the first work that capsule networks have
  been empirically investigated for  sequence-to-sequence problems.
  \item We  propose several techniques  to stabilize the dynamic routing process.
  We believe that these
  technique  should always be
  employed by capsule networks for the best performance.

\end{itemize}

\section{Linear Time Neural Machine Translation}
\begin{figure}[ht]
\begin{center}
\includegraphics[width=0.4\textwidth]{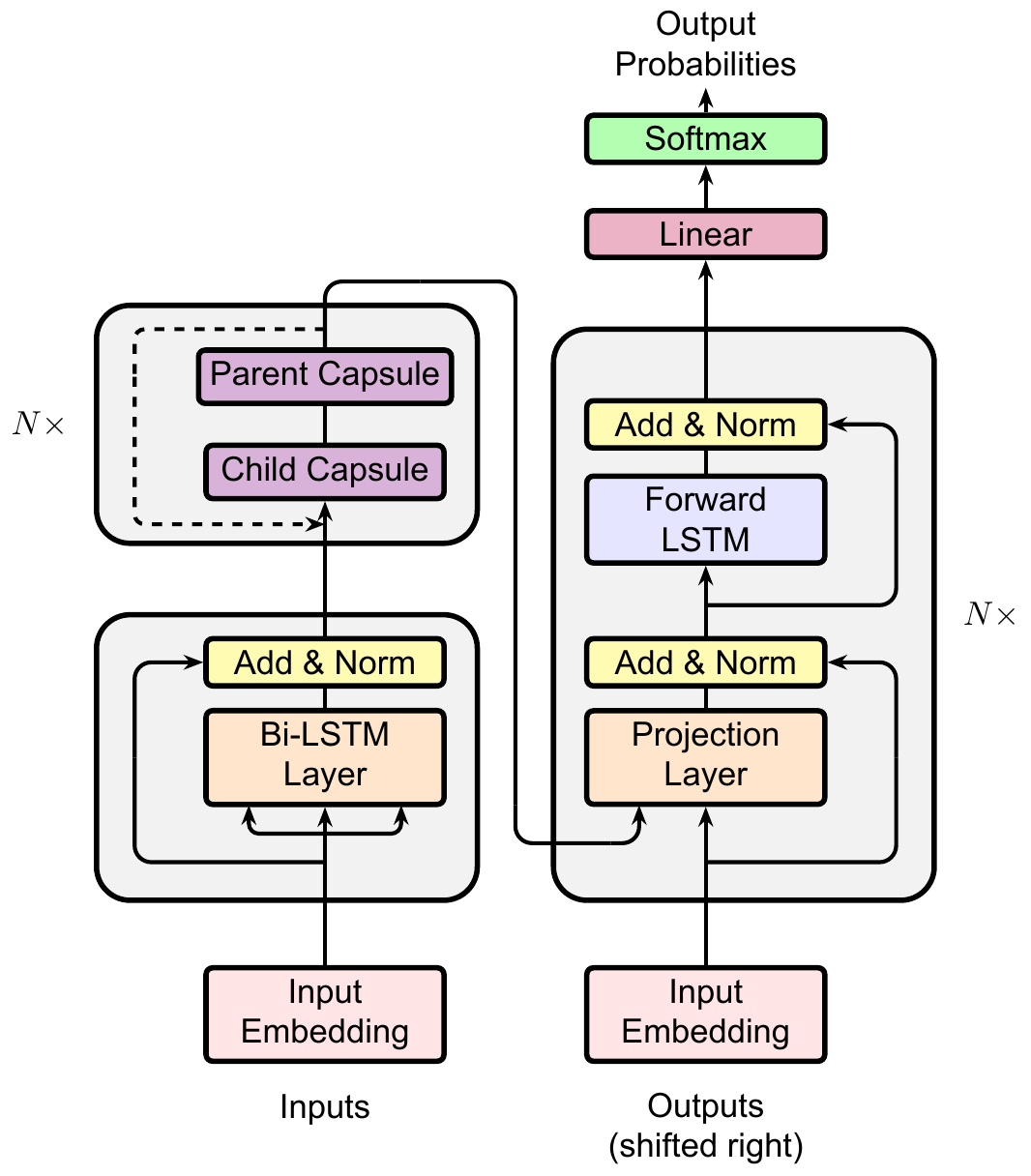}
\end{center}
\caption{\textsc{CapsNMT}:Linear time neural machine translation with capsule encoder}\label{f:overall}
\end{figure}
From the perspective of machine learning, the task of linear-time translation can be formalized
 as learning the conditional distribution $p(\y|\x)$ of a target sentence (translation) $\y$ given a source sentence $\x$.
 Figure~\ref{f:overall} gives the framework of our proposed linear time NMT with capsule encoder,
 which mainly consists of two components:
 a constant encoder that represents the input source sentence with a fixed-number of vectors,
 and a decoder which leverages these vectors to generate the target-side translation.
 Please note that due to the fixed-dimension representation of encoder, the time complexity of our model could be linear in the number of source words.

\paragraph{Constant Encoder with Aggregation Layers}
Given the input source sentence
$\x=x_1,x_2\cdots,x_L$,
Then, we introduce capsule networks to transfer $$ \X = [\x_1,\x_2,\cdots,\x_L]\in \bbR^{L\times d_x}.$$
The goal of the constant encoder is  to transfer the inputs $\X\in \bbR^{L\times d_v}$ into a pre-determined size
representation
$$\C=[\c_1,\c_2,\cdots,\c_M] \in \bbR^{M\times d_c}.$$
where
$M<L$ is the pre-determined size of the encoder output,
and $d_c$ is the dimension of the hidden states.

We first introduce a bi-directional
LSTM (BiLSTM) as the primary-capsule layer to incorporate forward and backward context information of
the input sentence:
\begin{equation}
  \begin{split}
    \overrightarrow{\h_t} &= \overrightarrow{\mathbf{LSTM}}(\h_{t-1},\x_t) \\
    \overleftarrow{\h_t} &= \overleftarrow{\mathbf{LSTM}}(\h_{t+1},\x_t) \\
    \h_t  &= [\overrightarrow{\h_t},\overleftarrow{\h_t}]
\end{split}
\end{equation}
Here we produce the sentence-level encoding $\h_t$ of  word $\x_t$ by concatenating the
forward $\overrightarrow{\h_t}$
and backward output vector $\overleftarrow{\h_t}$.
Thus, the output of BiLSTM encoder are a sequence of vectors
$\HH=[\h_1,\h_2,\cdots,\h_L]$ corresponding to the input sequence.

On the top of the BiLSTM,
we introduce several aggregation layers to map the
variable-length inputs into the compressed representation.
To demonstrate the effectiveness of our encoder,
we compare it with a more powerful aggregation method,
which mainly involves Max and Average pooling.

Both Max and Average pooling are the simplest ways to aggregate information,
which require no additional parameters while being computationally efficient.
Using these two operations,
we perform pooling along the time step as follows:
\begin{eqnarray}
  \h^{max}&=&\max([\h_1,\h_2,\cdots,\h_L]) \\
  \h^{avg}&=&\frac{1}{L}\sum^{L}_{i=1}\h_i
\end{eqnarray}
Moreover, because the last time step state $h_L$ and the first time step state $h_1$ provide complimentary information,
we also exploit them to enrich the final output representation of encoder.
Formally, the output representation of encoder consists of four vectors,
\begin{equation}
  \C=[\h^{max},\h^{avg},\h_1,\h_L].
\end{equation}
The last time step state $\h_L$ and
the first time step state $\h_1$ provide complimentary information, thus improve the performance.
The compressed  represention $\C$ is fixed for the subsequent translation generation,
 therefore
its quality directly affects the success  of building the Enc-Dec paradigm.
In this work,
we mainly focus on how to introduce aggregation layers with capsule networks to accurately produce the compressed representation of input sentences,
of which details will be provided in Section ~\ref{sec:capsule}.

\paragraph{LSTM Decoder}
The goal of the LSTM is to estimate the conditional probability $p(y_{t}|y_{<t}; x_1,x_2,\cdots, x_T)$, where $(x_1,x_2,\cdots, x_T)$ is the input sequence and $(y_1,y_2,\cdots, y_{T'})$ is its corresponding output sequence.

A simple strategy for general sequence learning is to map the input sequence to a fixed-sized vector, and then to map the vector to the target sequence with a conditional LSTM decoder\cite{sutskever2014sequence}:
\begin{equation}
  \begin{split}
    \sss_t &= \mathbf{LSTM}(\sss_{t-1}, \uu_{t}) \\
    \uu_{t} &= \W_{c}[\c_1,\cdots,\c_M] + \y_t
  \end{split}
\end{equation}
where $\y_t$ is the target word embedding of $y_t$,  $\uu_{t}$ is the inputs of LSTM at time step $t$,
$[\c_1,\cdots,\c_M]$ is the concatenation of the source sentence represention and $\W_c$ is the projection matrix.
Since $\W_{c}[\c_1,\cdots,\c_M]$ is calculated  in advance, the decoding time could be linear in the length of the sentence length.
At inference
stage, we only utilize the top-most hidden
states $\sss_{t}$ to make the final prediction with a
softmax layer:
\begin{equation}
  p(y_t|y_{t-1}, \x)=\mathbf{softmax}(\W_o\sss_t).
\end{equation}

Similar as ~\cite{vaswani2017attention}, we also employ a residual connection ~\cite{he2016deep} around each of
the sub-layers, followed by layer normalization~\cite{ba2016layer}.

\section{Aggregation layers with Capsule Networks}
\label{sec:capsule}
\begin{figure}[ht]
\begin{center}
\includegraphics[width=0.5\textwidth]{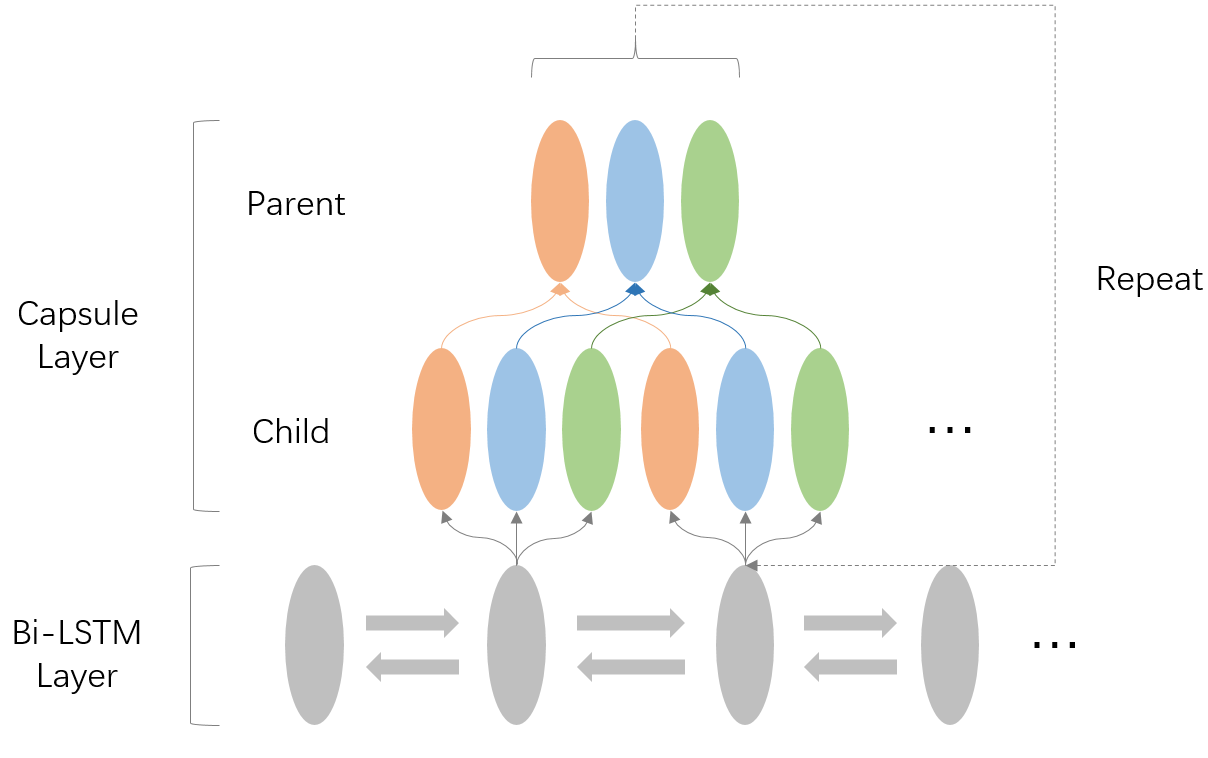}
\end{center}
\caption{Capsule encoder with dynamic routing by agreement}\label{f:capsencoder}
\end{figure}
The aggregation layers with capsule networks play a crucial role in our model.
As shown in Figure ~\ref{f:capsencoder},
unlike the traditional linear-time approaches collecting information in a bottom-up approach without considering the state of the whole encoding,
our capsule layer is able to iteratively decide the information flow,
where the part-whole relationship of the source sentence can be effectively captured.

\subsection{Child-Parent Relationships}
To compress the input information into the representation with pre-determined size,
the central issue we should address is to
determine  the information flow from the input capsules to the output capsules.

Capsule network is an ideal solution which is able t to address the representational
limitation and exponential inefficiencies of
the simple aggregation pooling method. It allows
the networks to automatically learn child-parent
(or part-whole) relationships.
Formally, $\uu_{i\rightarrow j}$ denotes the information
be transferred from the child capsule $\h_i$ into the parent capsule $\c_j$:
\begin{equation}
  \uu_{i\rightarrow j} = \alpha_{ij}\f(\h_i,\btheta_j)\label{eq:c1}
\end{equation}
where $\alpha_{ij}\in \bbR$ can be viewed as the
voting weight on the information flow from child
capsule to the parent capsule;
$\f(\h_i,\btheta_j)$ is the transformation function and in this paper, we use a single layer feed forward neural networks:
\begin{equation}
  \f(\h_i,\btheta_j) = \mathbf{ReLU}(\h_i\W_{j})\label{eq:c2}
\end{equation}
where $\W_{j}\in \bbR^{d_c\times dc}$ is the
transformation matrix corresponding to the position
$j$ of the parent capsule.

Finally, the parent capsule  aggregates all the incoming
messages from all the child capsules:
\begin{equation}
  \vv_i = \sum_{j=1}^{L} \uu_{j\rightarrow i}\label{eq:c3}
\end{equation}
and then squashes $\vv_i$ to $||\vv_i|| \in (0,1)$ confine.
ReLU or similar non linearity functions work well with single neurons. However we find
 that this squashing function works best with capsules.
 This tries to squash the length of output vector of a capsule. It squashes to 0 if it is a small vector and tries to limit the output vector to 1 if the vector is long.
\begin{equation}
  \begin{split}
    \c_i &= \mathbf{squash}(\vv_i) \\
        &= \frac{||\vv_i||^2}{1+||\vv_i||^2}\frac{\vv_i}{||\vv_i||}\label{eq:c4}
  \end{split}
\end{equation}

\subsection{Dynamic Routing by Agreement}
The dynamic routing process is implemented via an EM iterative process of refining
the coupling coefficient $\alpha_{ij}$ ,
which measures proportionally how much information is to be transferred
from $\h_i$ to $\c_j$.

At iteration $t$, the coupling coefficient $\alpha_{ij}$ is computed by
\begin{equation}
    \alpha_{ij}^{t} = \frac{\exp(b_{ij}^{t-1})}{\sum_k\exp(b_{ik}^{t-1})}
\label{eq:alpha}
\end{equation}
where $\sum \alpha_{j} \alpha{ij}$. This ensures that all the information from the child capsule $\h_i$ will be transferred to the parent.

\begin{equation}
    b^{t}_{ij} = b^{t-1}_{ij}+\c^{t}_j\cdot\f(\h_i^{t},\btheta^{t}_j)
\label{eq:b}
\end{equation}
This coefficient $b^t_{ij}$ is simply a temporary value
that will be iteratively updated with the value $b^{t-1}_{ij}$
of the previous iteration and the scalar product of $\c^{t-1}_j$ and $\f(\h^{t-1}_j,\btheta^{t-1}_j)$,
which is essentially the similarity between the input to the capsule and the output from the capsule.
Likewise, remember from above, the lower level capsule will send its output to the higher level capsule with similar output.
Particularly, $b^0_{ij}$ is initialized with 0.
The coefficient depends on the location and type of both the child and the parent capsules,
which iteratively refinement of $b_{ij}$.
 The capsule network can increase or decrease the connection strength by
 dynamic routing.
 It is more effective than the previous routing strategies such as max-pooling
 which essentially detects whether a feature is present in any position of the text,
 but loses spatial information
 about the feature.

 \begin{algorithm}[!ht]
  \caption{Dynamic Routing Algorithm}
  \begin{algorithmic}[1]
  \Procedure{Routing}{$[\h_1,\h_2,\cdots,\h_L]$,$T$}
  \State Initialize $\b^0_{ij}\leftarrow 0$
  \For{each $t \in \text{range}(0:T)$}
  \State Compute the routing coefficients $\alpha^t_{ij}$ for all $i\in[1,L], j\in[1,M]$ \Comment{From Eq.(\ref{eq:alpha})}
  \State Update all the output capsule $\c^t_j$ for all $j\in[1,M]$ \Comment{From
   Eq.(\ref{eq:c1},\ref{eq:c2},\ref{eq:c3},\ref{eq:c4})}
  \State Update all the coefficient $\b^t_{ij}$ for all $i\in[1,L], j\in[1,M]$ \Comment{From Eq.(\ref{eq:b})}
  \EndFor \\
  \Return $[\c_1,\c_2,\cdots,\c_M]$
  \EndProcedure
  \end{algorithmic}
 \end{algorithm}

 When an output capsule $\c_j^t$ receives the incoming messages $\uu^t_{i\rightarrow j}$, its state will be updated and the coefficient
 $\alpha_{ij}^{t}$ is also re-computed for the input capsule $\h^{t-1}_i$.
 Thus, we iteratively refine the route of information flowing, towards an instance dependent and context aware encoding of a sequence.
 After the source input is encoded into $M$ capsules, we map these capsules into vector representation by simply concatenating all capsules:
 \begin{equation}
   \C=[\c_1,\c_2,\cdots,\c_M]
 \end{equation}
 Finally matrix $\C$  will then be fed to the final end to end NMT model as
 the source sentence encoder.

 In this work, we also explore three strategies to
 improve the accuracy of the routing process.
\paragraph{Position-aware Routing strategy}
The routing process  iteratively decides what and
how much information is to be sent to the final
encoding with considering  the state of both the
final outputs capsule and the inputs capsule.
In order to fully exploit the order of the child and parent capsules to capture the child-parent relationship more efficiently,
we add ``positional encoding'' to the representations of child and parent capsules.
In this way, some information can be injected according to the relative or absolute position of capsules in the sequence.
In this aspect,
there have many choices of positional encodings proposed in many NLP tasks \cite{gehring2017convolutional,vaswani2017attention}.
Their experimental results strongly demonstrate that adding positional information in the text is more effective than in image since there is some sequential information in the sentence.
In this work, we follow ~\cite{vaswani2017attention} to apply sine and cosine functions of different frequencies.
\paragraph{Non-sharing Weight Strategy}
Besides, we explore two  types
of transformation matrices to generate the message
vector $\uu_{i\rightarrow j}$ which has been previously mentioned  Eq.(\ref{eq:c1},\ref{eq:c2}).
The first one shares parameters  $\btheta$ across different
iterations.
In the second design, we replace the shared
parameters with the non-shared strategy $\btheta^t$ where $t$
is the iteration step during the dynamic process.
We will compare the effects of these two strategies in our experiments.
In our preliminary, we found that non-shared weight strategy works slightly
better than the shared one which is in consistent with \cite{liao2016bridging}.

\paragraph{Separable Composition and Scoring strategy}
 The most important idea behind capsule networks is to measure the input and output similarity.  It is often modeled
 as a dot product function between the input and the output   capsule,
  and  the routing coefficient is updated correspondingly.
 Traditional capsule networks often resort to
 a straightforward  strategy
 in which the ``fusion'' decisions (e.g., deciding the voting weight
) are made based on the values of feature-maps. This is essentially a soft
template matching \cite{lawrence1997face}, which works for tasks like classification, however, is undesired for
maintaining the composition functionality of capsules. Here, we propose to employ separate
functional networks to release the scoring duty, and let $\btheta$ defined in Eq.(\ref{eq:c1}) be responsible for  composition.
More specifically, we redefined the iteratively scoring function in Eq.(\ref{eq:b}) as follow,
\begin{equation}
    b^{t}_{ij} = b^{t-1}_{ij}+\g(\c^{t}_j)\cdot\g(\h_i^{t}).
\label{eq:bfix}
\end{equation}
Here $\g(\cdot)$ is a fully
connected feed-forward network, which consists of two linear transformations with a ReLU activation and
is applied to each position separately.

\section{Experiments}
\subsection{Datasets}
We mainly evaluated \textsc{CapsNMT} on the widely used WMT English-German and
English-French
translation task.
The evaluation metric is BLEU.
We tokenized the reference and evaluated the
performance with multi-bleu.pl.
The metrics are exactly the same as in previous work
\cite{papineni2002bleu}.

For English-German, to compare with the
results reported by previous work, we used the same subset
of the WMT 2014 training corpus that contains
4.5M sentence pairs with 91M English words and
87M German words. The concatenation of news-test
2012 and news-test 2013 is used as the validation
set and news-test 2014 as the test set.

To evaluate at scale, we also report the results of
English-French.
To compare with the results
reported by previous work on end-to-end NMT,
we used the same subset of the WMT 2014 training corpus
that contains 36M sentence pairs. The concatenation
of news-test 2012 and news-test 2013
serves as the validation set and news-test 2014 as
the test set.
\begin{table*}[!ht]
\begin{center}
\begin{tabular}{l|l|c|c|c}

SYSTEM & Architecture &Time& EN-Fr BLEU &EN-DE BLEU \\
\hline
Buck et al. ~\shortcite{buck2014n} & Winning WMT14  &-& 35.7 & 20.7 \\
\hline
\hline
\multicolumn{5}{c}{Existing Enc-Dec Att NMT systems} \\
\hline
Wu et al. ~\shortcite{wu2016google} & GNMT + Ensemble &$|S|$$|T|$ & 40.4 & 26.3 \\
Gehring et al.~\shortcite{DBLP:journals/corr/GehringAGYD17} & ConvS2S &$|S|$$|T|$& 40.5 & 25.2 \\
Vaswani et al. ~\shortcite{vaswani2017attention} & Transformer (base)  &$|S|$$|T|$+$|T|$$|T|$& 38.1 & 27.3 \\
Vaswani et al. ~\shortcite{vaswani2017attention} & Transformer (large) &$|S|$$|T|$+$|T|$$|T|$& 41.0 & 27.9 \\
\hline
\hline
\multicolumn{5}{c}{Existing Enc-Dec NMT systems} \\
\hline
Luong et al. ~\shortcite{luong2015effective} & Reverse  Enc-Dec&$|S|$+$|T|$ & - & 14.0 \\
Sutskever et al. ~\shortcite{sutskever2014sequence} & Reverse stack  Enc-Dec &$|S|$+$|T|$& 30.6 &- \\
Zhou et al. ~\shortcite{zhou2016deep} & Deep Enc-Dec &$|S|$+$|T|$ & 36.3 & 20.6 \\
Kalchbrenner et al. ~\shortcite{kalchbrenner2016neural} & ByteNet& c$|S|$+c$|T|$ & - & 23.7 \\
\hline
\hline
\multicolumn{5}{c}{\textsc{CapsNMT} systems} \\
\hline
Base Model & Simple Aggregation & c$|S|$+c$|T|$ & 37.1 & 21.3 \\
Base Model & \textsc{CapsNMT} & c$|S|$+c$|T|$ & 39.6 & 25.7 \\
Big Model & \textsc{CapsNMT} & c$|S|$+c$|T|$ & \textbf{40.0} & \textbf{26.4} \\
\hline
Base Model & Simple Aggregation + KD & c$|S|$+c$|T|$ & 38.6 & 23.4 \\
Base Model & \textsc{CapsNMT} + KD & c$|S|$+c$|T|$ & 40.4 & 26.9 \\
Big Model & \textsc{CapsNMT}+ KD & c$|S|$+c$|T|$ & \textbf{40.6} & \textbf{27.6} \\
\end{tabular}

\end{center}
\caption{\label{tab:GEEXP} Case-sensitive BLEU scores on English-German and English-French translation. KD
indicates knowledge distillation~\cite{kim2016sequence}.
}
\end{table*}
\subsection{Training details}
Our training procedure and hyper parameter
choices are similar to those used by~\cite{vaswani2017attention}.
In more details,
For English-German translation and
English-French translation, we  use $50K$ sub-word tokens as vocabulary based on Byte Pair Encoding\cite{DBLP:journals/corr/SennrichHB15}.
We batched sentence pairs by approximate length, and limited input and output tokens per batch to $4096$ per GPU.

During training, we employed label smoothing of value $ \epsilon= 0.1$\cite{pereyra2017regularizing}.
We used a beam width of $8$ and  length penalty $\alpha=0.8$ in all the experiments.
The dropout rate was set to $0.3$ for the English-German task and $0.1$ for the English-French task.
Except when otherwise mentioned,
NMT systems had $4$ layers encoders followed by a capsule layer and $3$ layers decoders.
We trained for 300,000 steps on 8 M40 GPUs, and averaged the last $50$ checkpoints, saved at $30$ minute intervals.
For our base model, the dimensions of all the hidden states were set to $512$ and
for the big model, the dimensions were set to $1024$. The capsule number is set to $6$.

Sequence-level knowledge distillation is applied to alleviate multimodality in the training
dataset, using the state-of-the-art transformer base models as the teachers\cite{kim2016sequence}.  We decode the entire training set once
using the teacher to create a new training dataset
for its respective student.

\subsection{Results on English-German and English-French Translation}

The results on English-German and English-French translation
are presented in Table~\ref{tab:GEEXP}.
We compare \textsc{CapsNMT} with various other systems
including the winning system in WMT'14~\cite{buck2014n}, a phrase-based system whose language
models were trained on a huge monolingual text,
the Common Crawl corpus. For Enc-Dec Att
systems, to the best of our knowledge, GNMT is the best  RNN  based NMT system.
Transformer ~\cite{vaswani2017attention} is currently the SOTA system which is  about $2.0$ BLEU points better than GNMT on the English-German task and $0.6$ BLEU points better than GNMT on the English-French task.
For Enc-Dec NMT, ByteNet is the previous state-of-the-art system which has 150
convolutional encoder layers and 150 convolutional decoder layers.

On the English-to-German task, our big \textsc{CapsNMT} achieves the  highest
BLEU score among all the
Enc-Dec approaches which even outperform ByteNet, a relative strong competitor, by  $+3.9$ BLEU score.
In the case of the larger English-French task, we
achieves the  comparable
BLEU score among all the
with Big Transform, a relative strong competitor with a gap of only  $0.4$ BLEU score.
To show the power of the capsule encoder, we also make a comparison with
the simple aggregation version of the Enc-Dec model, and again
yields a gain of  $+1.8$ BLEU score on English-German task and $+3.5$
BLEU score on English-French task for the base model.
The improvements is in consistent with our intuition that
the dynamic routing policy is  more effective than the simple aggregation method.
It is also worth noting that for the small model, the capsule encoder approach
get an improvement of $+2.3$ BLEU score over the Base Transform  approach on
English-French task. Knowledge also helps a lot to bridge the performance
gap between \textsc{CapsNMT} and the state-of-the-art transformer model.

The first column indicates the time complexity of the network as a function of the length of the sequences and is denoted by Time.
The ByteNet, the RNN Encoder-Decoder are the only networks that have linear running time (up to the constant c).
The RNN Enc-Dec, however, does not preserve the source sequence resolution, a feature that aggravates learning for long sequences.
The Enc-Dec Att do preserve the resolution, but at a cost of a quadratic running time.
The ByteNet overcomes these problem with the convolutional neural network, however
the architecture must be deep enough to capture the global information of a sentence.
The capsule encoder makes use of the dynamic routing policy  to automatically learn the part-whole relationship and encode the source sentence into fixed size representation.
With the capsule encoder, \textsc{CapsNMT} keeps linear running time and
the constant $c$ is the capsule number which is set  to $6$ in our mainly experiments.

\subsection{Ablation Experiments}
In this section, we evaluate the importance of our
main techniques for training \textsc{CapsNMT}.
We believe that these
techniques are universally applicable across different
NLP tasks, and should always be
employed by capsule networks for best performance.
\begin{table}[!h]
\begin{center}
\begin{tabular}{l|c}
Model & BLEU\\
\hline
\hline
Base \textsc{CapsNMT} & $24.7$ \\
+ Non-weight sharing  & $25.1$ \\
+ Position-aware Routing Policy & $25.3$ \\
+ Separable Composition and Scoring & $25.7$ \\
\hline
+Knowledge Distillation  & $26.9$ \\
\end{tabular}
\end{center}
\caption{\label{tab:ABEXP}
English-German task: Ablation experiments of different technologies.
}
\end{table}
From Table~\ref{tab:ABEXP} we draw the following conclusions:
\begin{itemize}
  \item \textbf{Non-weight sharing strategy} We observed that the non-weight sharing strategy improves
  the baseline model
 leading to an  increase of  $0.4$ BLEU.
 \item \textbf{Position-aware Routing strategy} Adding the position embedding to the
 child capsule and the parent capsule can obtain an improvement of $0.2$ BLEU score.
 \item \textbf{Separable Composition and Scoring strategy} Redefinition of the dot product function
 contributes significantly to the quality of the model, resulting in an increase of $0.4$ BLEU score.
 \item \textbf{Knowledge Distillation}  Sequence-level knowledge distillation can still achive an increase of
 $+1.2$ BLEU.
\end{itemize}

\subsection{Model Analysis}
\begin{figure}[!ht]
\begin{center}
\includegraphics[width=0.4\textwidth]{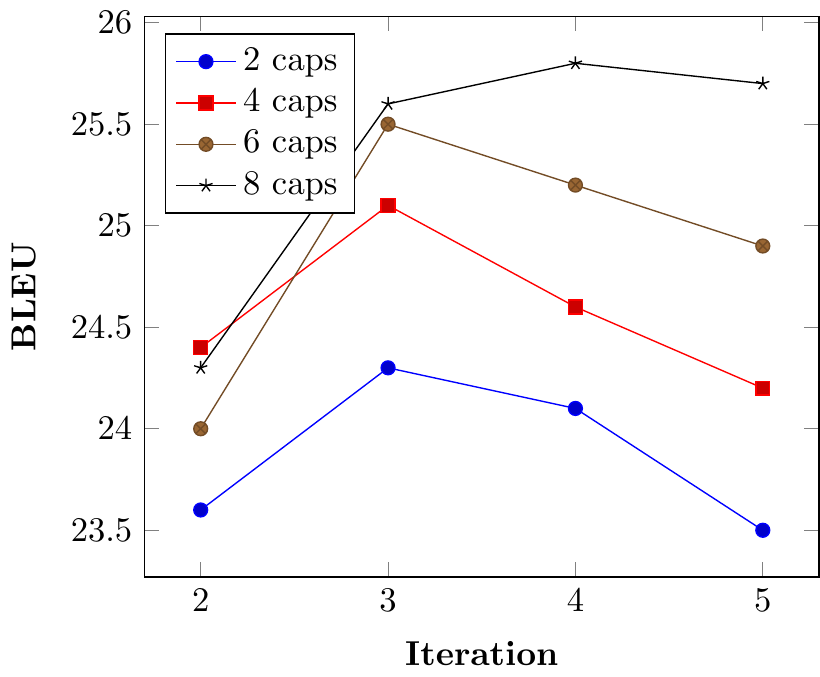}
\end{center}
\caption{Effects of Iterative Routing with different capsule numbers}\label{f:routing}
\end{figure}

In this section,  We study the attribution of \textsc{CapsNMT}.
\paragraph{Effects of Iterative Routing}
We also study how the iteration number affect the performance of aggregation on the English-German
task. Figure ~\ref{f:routing} shows the comparison of $2-4$ iterations in the dynamic routing process. The capsule number is set to $2,4,6$ and $8$ for each comparison respectively. We found that the performances on several different capsule number setting reach the best when iteration is set to 3. The results indicate the dynamic routing is contributing to improve the performance and a larger capsule number often leads to
better results.
\paragraph{Analysis on Decoding Speed}
\begin{table}[!h]
\begin{center}
\begin{tabular}{l|l|c}
Model & Num &Latency(ms) \\
\hline
\hline
Transformer & - & $225$ \\
\textsc{CapsNMT}& $4$ & $146$ \\
\textsc{CapsNMT}& $6$ & $153$ \\
\textsc{CapsNMT}& $8$ & $168$ \\
\end{tabular}
\end{center}
\caption{\label{tab:SPEEDEXP}
Time required for  decoding with the base model.
Num indicates the capsule number.
Decoding indicates the amount of time in millisecond required for
translating one sentence, which is averaged over
the whole English-German newstest2014 dataset.
}
\end{table}
We show the  decoding speed of
both the transformer and \textsc{CapsNMT} in Table~\ref{tab:SPEEDEXP}.
Thg results empirically demonstrates that \textsc{CapsNMT} can improve
the decoding speed of the transformer approach by $+50\%$.

\begin{figure}[!ht]
\begin{center}
\includegraphics[width=0.4\textwidth]{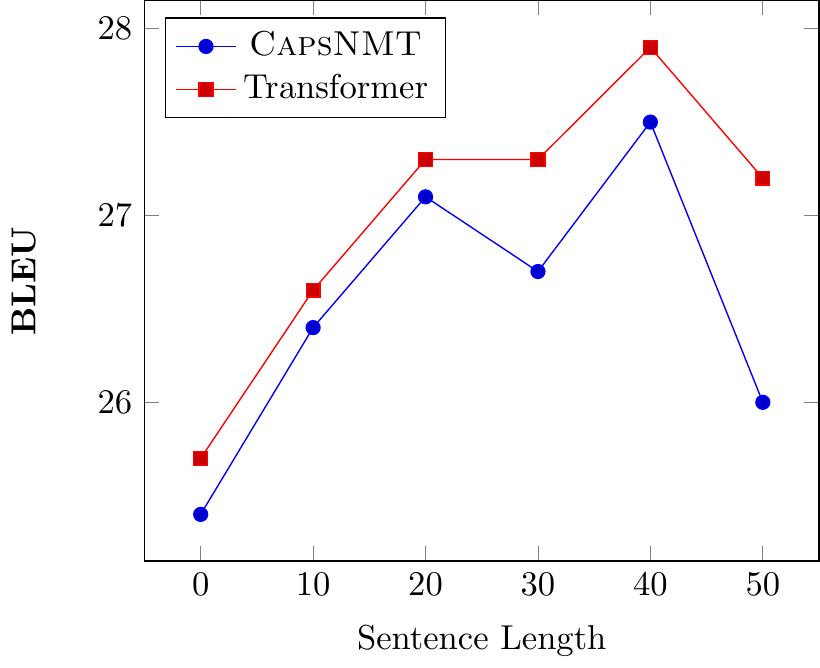}
\end{center}
\caption{The plot shows the performance of our system as a function of sentence length, where the
x-axis corresponds to the test sentences sorted by their length.}\label{f:bleu}
\end{figure}

\paragraph{Performance on long sentences}
A more detailed comparison between \textsc{CapsNMT} and Transformer can be seen in
Figure~\ref{f:bleu}.
In particular, we test the BLEU scores on sentences longer than $\{0, 10, 20, 30, 40,50\}$.
We were surprised to discover that the capsule encoder did well on medium-length sentences.
There is no degradation on sentences with less than 40 words, however,  there is still a gap
on the longest sentences. A deeper capsule encoder potentially helps to address the degradation
problem and
we will leave this  in the future work.

\paragraph{Visualization}
\begin{table}[!h]
\begin{center}
\begin{tabular}{l}
\footnotesize{
\textcolor{blue!100}{Orlando} \textcolor{blue!100}{Bloom} \textcolor{blue!10}{and} \textcolor{blue!10}{Miranda} \textcolor{blue!10}{Kerr} \textcolor{blue!10}{still} \textcolor{blue!10}{love} \textcolor{blue!10}{each} \textcolor{blue!10}{other}}\\

\footnotesize{\textcolor{blue!10}{Orlando} \textcolor{blue!10}{Bloom} \textcolor{blue!100}{and} \textcolor{blue!100}{Miranda} \textcolor{blue!10}{Kerr} \textcolor{blue!10}{still} \textcolor{blue!10}{love} \textcolor{blue!10}{each} \textcolor{blue!10}{other}}\\

\footnotesize{\textcolor{blue!10}{Orlando} \textcolor{blue!10}{Bloom} \textcolor{blue!10}{and} \textcolor{blue!10}{Miranda} \textcolor{blue!100}{Kerr} \textcolor{blue!10}{still} \textcolor{blue!10}{love} \textcolor{blue!10}{each} \textcolor{blue!10}{other}}\\

\footnotesize{\textcolor{blue!10}{Orlando} \textcolor{blue!10}{Bloom} \textcolor{blue!10}{and} \textcolor{blue!10}{Miranda} \textcolor{blue!10}{Kerr} \textcolor{blue!100}{still} \textcolor{blue!100}{love} \textcolor{blue!100}{each} \textcolor{blue!100}{other}}
\end{tabular}
\end{center}
\caption{A visualization to show the perspective of a sentence from 4 different capsules at the third iteration.}\label{tab:Visualization}
\end{table}
We visualize how much information each child capsule sends to the parent capsules. As shown in Table
~\ref{tab:Visualization}, the color density of each word denotes the coefficient $\alpha_{ij}$ at
iteration 3 in ~Eq.(\ref{eq:alpha}).
At first iteration,  the $\alpha_{ij}$  follows a uniform distribution since $b_{ij}$ is initialized to 0, and $\alpha_{ij}$ then will be iteratively fitted with dynamic routing policy.
It is appealing to find that afte 3 iterations, the distribution of the voting weights$\alpha_{ij}$ will converge to
a sharp distribution and the values
will be very close to $0$ or $1$.
It is also worth mentioning that the capsule seems able to capture
some structure information. For example, the information of phrase \emph{still love each other} will be
sent to the same capsule. We will make further exploration in the future work.
\section{Related Work}
\paragraph{Linear Time Neural Machine Translation}
Several  papers have proposed to
use neural networks to directly learn the conditional distribution from a parallel corpus\cite{kalchbrenner2013recurrent,sutskever2014sequence,cho2014learning,kalchbrenner2016neural}.
In \cite{sutskever2014sequence}, an RNN was used to encode a source sentence
and starting from the last hidden state, to decode a
target sentence.
Different the RNN based approach, Kalchbrenner et al.,~\shortcite{kalchbrenner2016neural} propose ByteNet
which makes use of  the convolution networks to build the liner-time NMT system.
Unlike the previous
work,
the  \textsc{CapsNMT} encodes the source sentence with
an iterative process to
decide the credit attribution between nodes from lower and higher layers.
\paragraph{Capsule Networks for NLP}
Currently, much attention has been paid to how developing a sophisticated encoding models to capture
the long and short term dependency information in a sequence.
Gong et al.,~\shortcite{gong2018information}  propose an aggregation mechanism to obtain a fixed-size encoding with a dynamic
routing policy.
Zhao et al.,~\shortcite{zhao2018investigating}  explore capsule networks
with dynamic routing for multi-task learning and
achieve the best performance  on six text
classification benchmarks.
Wang et al.,~\shortcite{wang2018sentiment} propose RNN-Capsule, a capsule model based
on Recurrent Neural Network (RNN) for sentiment analysis.
To the best of our knowledge, \textsc{CapsNMT} is the first work that capsule networks have
been empirically investigated for  sequence-to-sequence problems.
\section{Conclusion}
We have introduced \textsc{CapsNMT}  for linear-time NMT.
Three strategies were proposed to boost the performance of dynamic routing process.
The emperical results show that \textsc{CapsNMT} can
achieve state-of-the-art results with better decoding latency
in several benchmark datasets.

\bibliography{linearNMT}

\bibliographystyle{acl_natbib}
\end{document}